\documentclass[conference]{IEEEtran}
\IEEEoverridecommandlockouts
\usepackage{cite}
\usepackage{amsmath,amssymb,amsfonts}
\usepackage{algorithmic}
\usepackage{graphicx}
\usepackage{textcomp}
\usepackage{xcolor}
\usepackage{hyperref}

\def\BibTeX{{\rm B\kern-.05em{\sc i\kern-.025em b}\kern-.08em
    T\kern-.1667em\lower.7ex\hbox{E}\kern-.125emX}}
\begin{document}

\title{Event-based vision on FPGAs -- a survey \\
\thanks{The work presented in this paper was supported by: the program ''Excellence initiative –- research university'' for the AGH University of Krakow.}
}

\author{\IEEEauthorblockN{Tomasz Kryjak, Senior Member IEEE}
\IEEEauthorblockA{\textit{Embedded Vision Systems Group}, 
\textit{AGH University of Krakow, Poland}\\
\texttt{tomasz.kryjak@agh.edu.pl}}
}

\maketitle

\begin{abstract}

In recent years there has been a growing interest in event cameras, i.e. vision sensors that record changes in illumination independently for each pixel.
This type of operation ensures that acquisition is possible in very adverse lighting conditions, both in low light and high dynamic range, and reduces average power consumption.
In addition, the independent operation of each pixel results in low latency, which is desirable for robotic solutions.
Nowadays, Field Programmable Gate Arrays (FPGAs), along with general-purpose processors (GPPs/CPUs) and programmable graphics processing units (GPUs), are popular architectures for implementing and accelerating computing tasks.
In particular, their usefulness in the embedded vision domain has been repeatedly demonstrated over the past 30 years, where they have enabled fast data processing (even in real-time) and energy efficiency.
Hence, the combination of event cameras and reconfigurable devices seems to be a good solution, especially in the context of energy-efficient real-time embedded systems.
This paper gives an overview of the most important works, where FPGAs have been used in different contexts to process event data.
It covers applications in the following areas: filtering, stereovision, optical flow, acceleration of AI-based algorithms (including spiking neural networks) for object classification, detection and tracking, and applications in robotics and inspection systems.
Current trends and challenges for such systems are also discussed.

\end{abstract}


\section{Introduction}

Event cameras, also referred to as dynamic vision sensors (DVS) or neuromorphic cameras, have gained considerable attention in recent years.
This is evidenced by the rapidly growing number of scientific publications, including at top computer vision and robotics conferences: CVPR, ICCV/ECCV, WACV, IROS, ICRA and in renowned journals such as IEEE TPAMI, IEEE TCSVT, IEEE RAL etc. 
The first more widely commercially implemented solutions are also emerging, e.g. improving the quality of photos taken with smartphones\footnote{\href{https://www.prophesee.ai/2023/02/27/prophesee-qualcomm-collaboration-snapdragon/}{https://www.prophesee.ai/2023/02/27/prophesee-qualcomm-collaboration-snapdragon/}}.


Event cameras, due to their properties (cf. Section \ref{sec:event_cameras}), are also an attractive sensor for mobile robot perception systems in the broad sense, as well as other solutions that can be referred to as embedded.
In such applications, a key element is the computing platform on which the algorithm is implemented.
It should have enough computing power to process the data in real-time with low latency and be energy efficient.
The vast majority of research published to date on event-based vision uses general-purpose processors (GPP/CPU) or programmable graphics cards (GPU) as the computational platform, especially in solutions using deep neural networks.
On the other hand, it has been known for more than 20 years that configurable devices are an attractive platform for the realisation of embedded vision systems: from classical to those using artificial intelligence methods \cite{Gandhare2019Survey, Shawahna_2019}.
They are characterised by relatively low power consumption, low latency, support parallelism and allow prototyping and multiple modifications of algorithms, also in target devices.
Hence, the use of FPGAs for event data processing seems to be a fairly natural direction for research and subsequent application development.


This article brings together the vast majority of research papers published between 2012 and the first half of 2024, which used FPGAs (Field Programmable Gate Arrays) or SoC FPGAs (System on Chip FPGAs) to process event data.
The following applications were distinguished: event data filtering (pre-processing), optical flow, stereovision, object classification, detection and tracking realised ``classically'', application of artificial intelligence algorithms (including spiking neural networks -- SNNs) and other applications.
Such a compilation allows interested readers to learn about the opportunities, advantages, challenges and drawbacks of using FPGAs for event data processing.
It also indicates the level of progress of the work in each area.
The most important contributions of the work can be summarised as:
\begin{itemize}    
    \item first review of scientific articles on the topic of event data processing using FPGAs and SoC FPGAs covering the period 2012-2024 (up to mid-year),
    \item a summary of the achievements to date and identification of areas where the application of SoC FPGA platforms has not been sufficiently explored (knowledge gaps).
\end{itemize}


The remainder of this paper is organised as follows:
Section~\ref{sec:event_cameras} briefly presents the technology of event cameras.
Section~\ref{sec:event_fpga} presents a synthetic overview of the work on event data processing in FPGAs.
The article concludes with a~discussion and a brief summary.


\section{Event cameras}
\label{sec:event_cameras}

Event cameras are inspired by the way the retina works, where photoreceptors (cones, rods) are activated by incident light, independently for each photosensitive element.
In the event camera, this process has been replicated -- each pixel acts independently (asynchronously) and if a change in brightness is recorded that exceeds a preset threshold, an event is generated in the form: \(E = \{x, y, t, p\}\), where $x$ and $y$ are the spatial coordinates of the pixel, $t$ is the timestamp and $p$~is the polarity (direction of brightness change).
It should be added that the monitoring of brightness changes is done on a~logarithmic scale and in the analogue part of a single pixel.


This mode of operation has a number of implications.
Unquestionable advantages include: 
(1) correct operation in adverse lighting conditions -- both in low light and in high dynamic range conditions (estimated at 120 dB compared to 50-60 dB for conventional frame cameras),
(2) less sensitivity to motion blur compared to frame cameras,
(3) low latency, defined as the time from the onset of a change in brightness to the appearance of this information at the sensor output,
(4) high temporal resolution -- timestamps are usually transmitted with an accuracy of 1 microsecond (1 MHz clock),
(5) no redundancy of information -- no changes in the scene, no events, resulting in lower bandwidth requirements for data transmission and higher energy efficiency (it should be noted that this applies to the ``average'' case -- for example, the occurrence of very dynamic movement or flashing lighting in the scene can lead to a data stream comparable to, or even greater than, that of frame cameras).


Challenges, on the other hand, include:
(1) the lack of absolute brightness information, which is essential in some scenarios (e.g. detection of static objects when the camera is also stationary),
(2) the relatively high level of noise, which makes it necessary to use some kind of filtering method,
(3) an unusual data format -- a sparse spatio-temporal cloud that is not directly compatible with existing processing methods for frame cameras and implies either the need to transform events into pseudo-images or to use less standard processing methods e.g. graph neural networks, spiking neural networks or grouping events into voxels.


In the context of data processing in reconfigurable systems, the last property should be particularly emphasised.
A typical frame camera data stream is very structured -- successive pixels are transmitted line by line and frame by frame.
This brilliantly facilitates the design of an efficient computational architecture and allows to take full advantage of computational parallelisation, including pipelining -- both for ``classical'' methods and deep convolutional neural networks. 
In the case of event cameras, the form of the data stream depends on the objects and the dynamics of the scene, which poses a major challenge.

For more information on event camera technology and its features and applications, refer to the very good review article \cite{gallego2020event} and a very recent review focused on automotive applications \cite{Shariff_2024}.

\section{Event-based vision on FPGA}
\label{sec:event_fpga}

\begin{figure}
    \centering
    \includegraphics[width=0.8\linewidth]{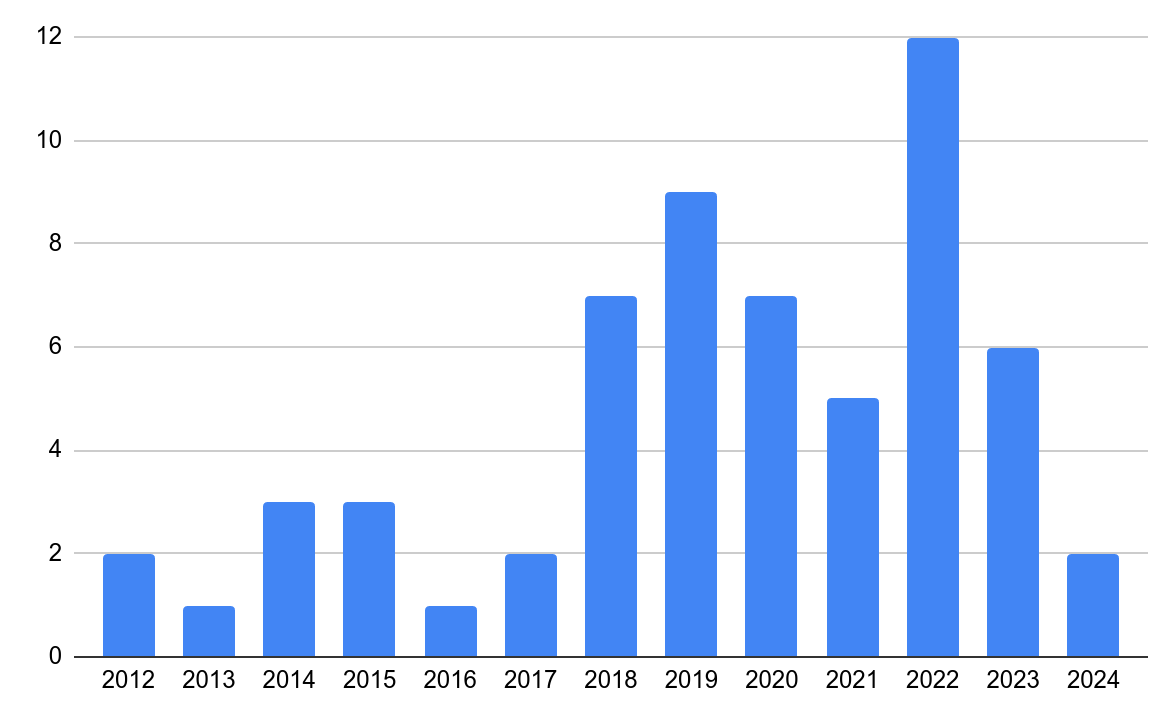}
    \caption{Number of publications per year.}
    \label{fig:years}
\end{figure}

The starting point for this review was a query for articles that use FPGAs to process event data.
This was done using a~combination of the following approaches: a search in Google Scholar and Scopus (mainly the keywords FPGA or VLSI and event camera and DVS), a list of articles maintained by Guillermo Gallego from TU Berlin\footnote{\url{https://docs.google.com/spreadsheets/d/1\_OBbSz10CkxXNDHQd-Mn\_ui3OmymMFvm-lW316uvxy8/edit\#gid=0}} and the Robotics and Perception Group from the University of Zurich\footnote{\url{https://github.com/uzh-rpg/event-based_vision_resources}}, and an analysis of the references given in the reviewed articles.
As a~result, 60 articles from the period 2012-2024 (first half) were collected.
Due to the relatively small size of the field, it is currently still possible to review the vast majority of papers.
However, despite best efforts, it is possible that a small number of papers have been omitted, particularly recent ones or these with less obvious titles or keywords.
In addition, pre-prints were not included in the review.
The number of articles per year is summarised in Figure \ref{fig:years}.


By analysing the collected articles, it is furthermore possible to distinguish research groups related to the considered topic.
Due to frequent international collaborations, it was decided to assign an article to a research group/country based on the first author's affiliation.
By far the largest number of papers came from the University of Seville (Robotic and Technology of Computer Laboratory\footnote{Due to possible changes in the names of research groups, this data should be treated with some uncertainty.}) -- at least 11 papers. 
In addition, 9~papers were published by a group associated with the Institute of Neuroinformatics of ETH Zurich, 5 by a group associated with the Instituto de Microelectrónica de Sevilla, University of Seville\footnote{It should also be noted that many papers were produced in collaboration between the University of Seville and ETH Zurich.}, and 4 by the Computer Architecture and System Research (CASR) group of the University of Hong-Kong, 3~by Temasek Laboratories, National University of Singapore, and 3~by the Embedded Vision Systems Group of AGH University of Krakow, Poland.
In the early 2012-2014 period, a group associated with AiT in Austria was also active in this field, publishing 3 papers.

\begin{figure}
    \centering
    \includegraphics[width=1\linewidth]{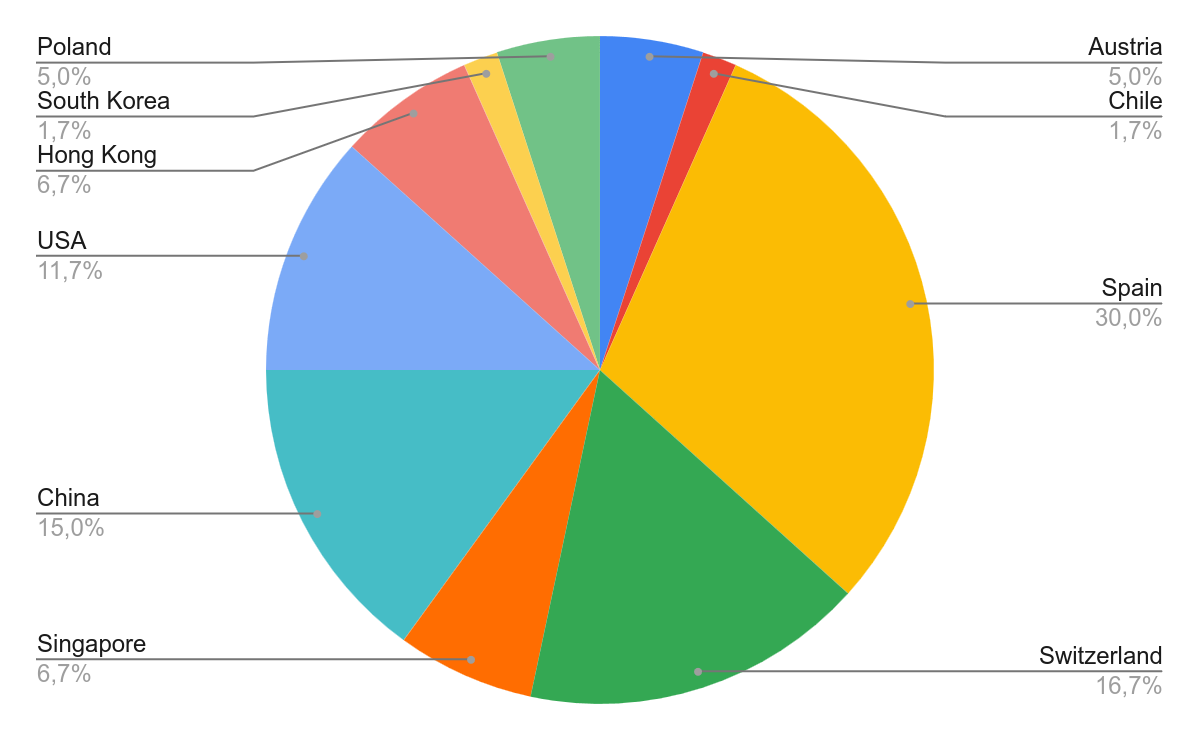}
    \caption{Publications by country (affiliation of the first author).}
    \label{fig:countries}
\end{figure}

A breakdown of publications by country is shown in Figure~\ref{fig:countries}.
Note that in the USA and China, teams published single papers, while in Spain, Switzerland, Hong-Kong, Singapore, Austria or Poland it is possible to identify teams that have published 3 or more papers. 

For the works discussed below, the algorithm implemented in hardware, the resolution of the event stream, the FPGA devices used (in curly brackets after the reference), the clock frequency and the event processing rate (MEPS -- Mega Events Per Second) or latency (if provided) are given in a~very synthetic form.
It is also indicated whether and how the solution has been evaluated, especially with a reference to previous works.
The type of publication is also indicated after the reference: (C) -- conference, (SC) -- short conference article/demonstration, (LC) -- long (more than 10 pages) conference article, (J) -- journal, (SJ) -- journal brief.


\subsection{Filtration}

Event data filtering is a very important issue, because -- due to transistor junction leakage and parasitic photo-currents -- event cameras are characterised by a high level of noise.
This is visible in the form of individual, uncorrelated events that are not related to the actual changes occurring: lighting or object and/or camera movement.


The first paper to address the implementation of filtering in FPGAs was \cite{Linares-Barranco_2015}(C)\{Lattice ECP3\}. 
It used a Background Activity Filter (BAF), which analyses the timestamps of incoming events and compares them with previously stored timestamps associated with correlated events.
On this basis, it is determined whether an event should be considered real or noise.
In addition, propagation of timestamps to neighbouring pixels was applied.
The solution uses a 32-bit word array of size 128$\times$128 to store the timestamps, which is implemented in BRAM resources.
As the proposed filtering is part of the tracking system, it has not been evaluated separately.
Information on maximum frequency and latency is also not provided.
The described timestamp analysis mechanism is the basis of most event data filtering methods.


A biologically-inspired algorithm called LDSI -- Less Data Same Information -- was proposed in the article \cite{Barrios_Aviles_2018}(J)\{Virtex 6\footnote{If the manufacturer of the FPGA is not mentioned, it should be assumed to be Xilinx.}\}.
Its task is not only to filter noise, but also to reduce the event stream while preserving the information contained.
The neuron model used is inspired by the bipolar cells of the retina.
The module works with a 128$\times$128 event camera, runs at 177 MHz clock and has a relatively low resource consumption (671 LUTs, 40 BRAMs).
The filtering method has not been formally evaluated, and there is no information on maximum throughput.


The paper \cite{Linares_Barranco_2019}(J)\{Spartan 6\} presents, among other things, an improved version of the BAF filter from \cite{Linares-Barranco_2015}.
The modification consists in a better consideration of the spatial context in the filtering process.
In addition, a simple mask filter dedicated to the removal of so-called hot or uninformative pixels has also been proposed.
It is built from two matrices -- the first one is used to detect pixels that frequently generate events and the second one is used to label and mask them.
The filtering has not been formally evaluated.
The solution worked for 128$\times$128 pixels DVS.


The paper \cite{Bisulco_2020}(C)\{Spartan 6\} proposes a filtering algorithm consisting of four stages: event parsing (omitting temporal information), coincidence detection (analysing the spatial co-occurrence of events), aggregation-subsampling (aggregating the occurrence of events over time) and Huffman encoding (consisting of 256 words).
The system works with a DVS sensor with a resolution of 480$\times$320 pixels and uses 598 FFs, 4053 LUTs, 1 DSP and 103 BRAMs.
The proposed filtering method has not been formally evaluated.


A filtering method referred to as HashHeat with a computational complexity of $O(C)$ was proposed in the paper \cite{Guo_2020}(C)\{Artix 7\}.
It does not require the storage of timestamps and its computational complexity does not depend on the spatial resolution of the input stream.
Its basis is a hash function operating in Euclidean space.
The solution operates at 100 MHz and the latency for a single event is 10 ns.
The filtering results were compared with several popular algorithms on the DVS Gesture set \cite{Amir_2017}.


In the work \cite{Khodamoradi_2021}(C)\{Artix 7\} it is assumed that, due to the sparse and random nature of noise, it is sufficient to store information about the occurrence of an event only in a given row or column.
This allows a computational complexity of $O(N)$ ($N$ -- spatial resolution of the sensor).
The authors conducted a theoretical performance analysis, as well as a~practical experiment comparing the proposed approach with two popular filtering methods with higher memory complexity.


In the paper \cite{Kowalczyk_2022}(C)\{Zynq US+ MPSoC\}, a method based on the analysis of timestamps is proposed. 
However, the timestamps are not substituted, but filtered (averaged) over time.
Evaluation of the algorithm has been carried out on a~number of different sequences, but without comparison with other methods.
The implementation has a low resource usage and a high throughput of 386.9 MEPS (Mega Events Per Second) and was designed to work with a 1280$\times$720 DVS.


As the most recent published work in this field should be considered \cite{Rios_Navarro_2023}(C)\{Zynq SoC, Zynq US+ MPSoC\}.
It proposes the MLPF (Multi Layer Perceptron Denoising Filter) algorithm.
It is the first solution using a neural network that is trained on labelled data.
A formal evaluation and comparison with other methods has been carried out for the method.
It has a very good filtering efficiency, however, a lower throughput of approximately 23 MEPS. 


In summary, the topic of event data filtering has so far been addressed in 8 scientific papers.
A promising direction seems to be the use of deep learing methods (as in the paper \cite{Rios_Navarro_2023}).
However, the challenge, as in any system with a neural network, is a computational complexity.
In addition, not in all articles do the authors carry out a formal evaluation of the proposed method.
There is also a lack of a single, established dataset on which filtering performance results can be reported.



\subsection{Optical flow}

Optical flow is the pattern of apparent motion of objects, surfaces, and edges in a visual scene caused by the relative motion between an observer and a scene.
Its analysis allows the detection of moving objects as well as the determination of the camera's ego-motion relative to its surroundings, which is used, for example, to stabilise multi-rotor UAVs.
A more extensive discussion of the issue can be found in the paper \cite{Zhai_2021}.
Optical flow is one of the fundamental concepts in computer vision, hence the attempt to determine it from the event camera data stream, also in FPGAs, seems to be an obvious research direction.


The first paper to address this topic was \cite{Liu_2017}(C)\{Spartan 6\}.
It proposed a method based on block matching, an approach known from classical computer vision.
In the first step, events are converted into binary pseudo-images (polarity is ignored), which are stored in BRAM -- two previous ones ($t_1$ and $t_2$) and the current one ($t_0$).
Then, for a new event from image t1, a 9$\times$9 context is taken and compared with 9 neighbouring contexts from t2.
The Hamming distance is used for matching.
For a 50 MHz clock, a single event processing time of 220 ns was achieved, which translates to 5 MEPS.
The evaluation used the well-known measures: AAE (Average Angular Error) and AEE (Average Endpoint Error), but it was not specified on which dataset.


A hardware implementation of an iterative plane-fitting algorithm to event-based optical flow estimation is presented in the paper \cite{Aung_2018}(C)\{Spartan 6\}. 
A processing rate of 2.75 MEPS was achieved, which is sufficient for a sensor with a resolution of 304$\times$240.
A ``rotating bar'' sequence, also considered in earlier works, was used for evaluation, but the paper lacks a~direct comparison with previous works.
The authors emphasise that the change to fixed-point arithmetic resulted in a slight decrease of performance.


In the paper \cite{Pivezhandi_2020}(C)\{Zynq US+ MPSoC\}, the authors focused on a histogram generation method for optical flow determination.
In the first step, events are stored in a buffer built from a bank of RAMs, with the data being compressed -- instead of absolute timestamps, only differences are stored. 
This is followed by background activity filtering and decompression of the timestamps.
Ultimately, this allows the histogram to be updated.
The solution allows processing of 200 MEPS.
No direct reference to optical flow determination is made in the paper.


In the paper \cite{Liu_2022}(J)\{Zynq SoC\}, the authors presented a~hardware implementation of the feature point (corner) detection algorithm referred to as SFAST (Slice-based FAST) and the calculation of block (sparse) optical flow for them -- ABMOF (Adaptive Block Matching Optical Flow).
It should be noted that a multi-scale method was used, which significantly improves efficiency.
The module is characterised by a high use of memory (BRAM) and DSP computational resources.
The dense flow calculation variant allows processing of 1~MEPS and the sparse flow (for feature points only) of more than 16 MEPS.
It is certainly worth noting the very extensive evaluation and comparison with two other optical flow computation methods for event data.
In addition, the corner detection method itself is presented in \cite{Liu_2019}(SC)\{Zynq SoC\}.


A module for optical flow determination using the fusion of event and vision data is presented in the article  \cite{Lele_2022}[C]\{Virtex US+\}.
For event data, a 3-layer Leaky CNN network is used, and for frames the classical Farneback algorithm.
The fusion process uses a confidence map, which is calculated from the event input.
The evaluation was performed on a set of MVSEC (sequences recorded from a camera mounted on a drone).
The authors estimate that 41 OF results (frames) per second will be possible.
However, in the current version, the Farneback module was not implemented in hardware.


In summary, the issue of implementing an OF module for event data in an FPGA has been addressed in 6 papers.
Most of the algorithms use different kinds of pseudo-images as representations.
The lack of an uniform dataset used for evaluation is noticeable.
Only one paper \cite{Lele_2022} addresses the fusion of event and vision data, using a neural network to determine the OF for events.


\subsection{Stereovision}

Alongside optical flow, depth estimation is also very important area of computer vision.
In addition, the algorithms performing this task are characterised by high computational complexity, which is a direct motivation for their implementation in FPGAs.

In the conference paper \cite{Humenberger_2012}(C)\{Spartan 3\}, a stereovision module was proposed as part of a fall detection system.
Incoming events (304$\times$240) are aggregated into pseudo-frames, for which a typical matching is then realised using a normalised SAD (Sum of Absolute Difference) in a 15$\times$15 window.
The paper lacks details of the FPGA implementation, only a reference to a previous work is provided.
There is also no formal evaluation of the stereovision module.
However, in the context of the whole system, an evaluation was performed on a~custom dataset with recorded falls.
It should be noted that after changing the representation from event data to pseudo-frames, the system architecture is similar to typical vision solutions.
More technical aspects of this solution have been presented in the paper \cite{Belbachir_2012}(C)\{Spartan 3\}, however, no information on performance or efficiency is available.


The paper \cite{Eibensteiner_2014}(C)\{Spartan 3\}, building on previous works \cite{Humenberger_2012} and \cite{Belbachir_2012}, presents an implementation of a stereovision system in an FPGA.
The authors divide the image into 16$\times$16 windows and perform matching only for windows that have a sufficiently high event activity level.
A rectification module and a disparity consistency check are also implemented.
For a~sensor with a resolution of 128$\times$128 and a 100 MHz clock, a~latency of 13.7 $\mu s$ was obtained, which the authors converted to 1140 fps (frames per second).
Evaluation was carried out on two of the authors' sequences and the comparison was limited to the SAD method from earlier work.


The article \cite{Camunas-Mesa_2014}(C)\{Spartan 6\} presents an implementation of event-driven convolution module arrays for a 128$\times$128 resolution camera.
The output generates, for example, Gabor filtering results that can be used in a stereovision algorithm.
However, implementation and evaluation details of this solution are missing.


In the article \cite{Domínguez_Morales_2019}(J)\{Virtex 5\}, a~stereovision system for a~128$\times$128 camera is presented.
In the first step, rectification was performed.
The description of the matching algorithm is not explicit, but it can be inferred that histogram representations are constructed on which the fitting is performed, taking into account the limitations of the epipolar geometry.
The method has not been formally evaluated.


The article \cite{Domínguez_Morales_2_2019}(J)\{Virtex 5\} focuses on the issue of calibrating the stereovision system.
Using a special calibration board, the authors obtained the projection matrices and, using them, recalculations were performed to reconstruct the actual X, Y, Z coordinates from the u, y projection coordinates of the left and right cameras.
The solution was tested in several cases -- both static and in motion.
The article provides little detail on the hardware implementation -- it processes data from a~128$\times$128 resolution camera, and the latency for a~single event is 10 ms.


The focus of the article \cite{Zhang_2021}(C)\{Zynq SoC\} is on the acceleration of rectification. 
It should be noted that this process does not differ from the computation for pixels acquired from a classical camera.
For data with a resolution of 400$\times$250, a~single module can recalculate up to 2200 frames per second (no information on the clock frequency is available).


The article \cite{Yu_2022}(J)\{Zynq SoC\}) is an extension of the paper \cite{Zhang_2021}.
It is worth noting that the authors do not use a typical event camera, but a spike camera that emits an event when the summed luminance intensity is greater than a preset threshold (analogy of the integrate-and-fire neuron).
For this device, they present a method of image reconstruction -- generating a kind of pseudo-image and the rectification already presented.
The calculation of the depth map itself is not realised.


The article \cite{li_2022}(C)\{Zynq SoC\} proposes an acceleration of the event-based monocular multi-view stereo (EMVS) approach, that is, depth reconstruction based on a recorded sequence whose motion trajectory is known.
The authors made some modifications to the EMVS method and then proposed acceleration of selected parts of the algorithm: event back-projection and volumetric ray counting.
The solution was clocked at 120 MHz and can process more than 1.8 MEPS.


The article \cite{Kim_2022}(J)\{Intel Cyclone IV\} presents a hardware implementation of a spiking cooperative network to determine stereo correspondences.
This method does not require the creation of an intermediate representation, but directly processes incoming events.
The module runs with a 50 MHz clock and has a latency of 2 ms for an input resolution of 128$\times$128.
Evaluation was carried out on a custom dataset.


To summarise, the majority of the papers use event representations, and only one paper \cite{Kim_2022} attempts to process event data ``directly''.
This is also the only work that uses neural networks.
As with optical flow, the lack of use of a uniform dataset for evaluation is noticeable.


\subsection{Object detection, recognition and tracking}

In this section, works using ``classical'' algorithms, i.e. not using deep networks, are summarised.
Research on object detection and recognition is presented first, followed by object tracking.


The paper \cite{Ramesh_2019}(C)\{Zynq SoC\} proposes a method for feature extraction from an event stream called PCA-RECT, which was used in an object detection and classification task.
In a first step, filtering is performed and then a dynamically updated activity count matrix is created for each pixel.
This is subjected to convolution (averaging) filtering.
A context is then selected from this matrix, which is projected to lower-dimensional subspace using PCA.
Based on this, a dictionary is created using the k-d tree nearest-neighbour search algorithm.
The SVM algorithm is used for classification and the activation map associated with the dictionary is used for location determination.
The authors also propose a variant without PCA, which is less computationally complex and better suited for implementation in an FPGA.
The modules described were run with a clock of 100 MHz (latency 550 ns).
Classification evaluation was performed on the N-MNIST and N-Caltech101 \cite{orchard2015converting} datasets, and detection on the author's dataset.
An extended version of the work was published in the journal \cite{Ramesh_2020}.
The evaluation was expanded, the description had been improved, and the possibility of classification at any time moment was added.


In the article \cite{He_2020}[J]\{Zynq SoC\}, a classification system for event data based on simple binary features and random ferns classifier is proposed.
Features are calculated by analysing the occurrence of positive and negative events in randomly selected areas for a certain fragment of the event stream.
The data is classified similarly to the Bayes method, using ferns that are determined from the features.
It should be noted that the method allows for a very simple implementation of online training.
The evaluation was carried out on the sets: MNIST-DVS, Poker-DVS and Posture-DVS. 
The module for the 100 MHz clock allows processing of 100 MEPS. 
Note that the resource usage grows exponentially with one of the feature vector parameters, which may be a limitation of the solution for more complex problems.


A hardware accelerator for the Hierarchy Of Time-Surfaces (HOTS) complex event data representation was presented in the paper \cite{Tapiador-Morales_2020}[J]\{Zynq SoC\}.
The module ran at a clock of 100 MHz and the reported throughput ranged from 0.16 MEPS to 2 MEPS. 
Evaluation was carried out for a gesture recognition application -- the authors recorded an event sequence with a resolution of 304$\times$240 pixels.
The computational quantisation used resulted in a decrease in performance of approximately 1 percentage point.


In paper \cite{Xu_2023}(LC)\{Zynq SoC\}, a~biologically-inspired stereovision obstacle detection system designed for a drone is proposed.
The system consists of several modules: filtering (using IMU sensor data and a stereo coherence condition), pseudo-image generation, and detection and localisation of matched features.
Computation was split between the processor system and the programmable logic of the SoC FPGA device.
Due to the unusual type of the system, the tests were carried out on the author's sequences (indoor and outdoor drone flight). 
Thus, the work does not include a direct comparison to other studies.


In the aforementioned paper \cite{Linares-Barranco_2015}, an object tracking module was proposed in addition to filtering.
It is based on the detection and tracking of clusters, i.e. areas where groups of events occur.
The module was tested in two scenarios: for slow and fast moving objects.


The paper \cite{Moeys_2016}(C)\{Spartan 6\} proposed a hardware implementation of the Retinal Ganglion Cell model which detects moving objects.
It is based on the Object Motion Cell (OMC), which can detect the presence of small moving objects.
This information can then be used for tracking.
For a 50 MHz clock, the latency was 220 or 440 ns depending on the number of OMC modules.


The paper \cite{Liu_2_2017}(C)\{Spartan 6\} presents a concept similar to that of the paper \cite{Moeys_2016}.
The Approach Sensitive Cell mechanism was implemented, whose task is to detect objects that approach the camera.
At a clock frequency of 50 MHz, a latency of 160 ns was achieved.
The results of this work are also described in more detail in a journal article \cite{Linares_Barranco_2018}, where, among other things, the application of the solution on a mobile robotic platform is shown.


In the paper \cite{Linares_Barranco_2019}(J)\{Spartan 6\}, the concept from \cite{Linares-Barranco_2015} was developed and combined with \cite{Moeys_2016}.
A library of four hardware modules: background activity filter, mask filter, cluster object tracker and object motion detector was proposed.
The implementation works with a DVS with a resolution of 128$\times$128 pixels.


The article \cite{Sengupta_2021}(C)\{Kintex 7\} presented an embedded system for event data processing.
In addition to the hardware modules, it included a RISC-V processor.
As an example application, filtering, corner detection, clustering and tracking were realised.
The system works with a 128$\times$128 pixel camera and can perform 19 MOPS (Mega Operations Per Second).


In the paper \cite{Gao_2022}(LC)\{Zynq SoC\}, the authors propose a~multiple object tracking (MOT) system using event cameras.
It is based on Attention Units that analyse incoming events in defined areas (one AU, one object) (realised in parallel in reconfigurable resources).
This process takes place under the supervision of a controller (implementation in a processor system).
The authors proposed 3 AU implementation variants with different ratios of computational complexity to efficiency.
Evaluation was carried out on 3 test sets.
An improved version of the system, with greater power-efficiency and enhanced performance, was presented in a journal publication \cite{Gao_2023}. 


A neural-inspired ultra-high-speed moving object filtering, detection, and tracking scheme is proposed in the paper \cite{Zhu_2022}(J)\{Zynq US+ MPSoC\}. 
Based on the analysis of the abstract\footnote{Full text not available in IEEE subscription.}, it can be inferred that object detection and tracking uses thresholding and connected component labelling.
The accelerator can process up to 20000 spike images with a~resolution of 250$\times$400 per second.


The paper \cite{Shi_2022}(C)\{Zynq US+ MPSoC\} presents an object tracking system consisting of four modules: denoising (Double Window Filter), corner detection (Slice-based FAST), optical flow (ABMOF \cite{Liu_2022}) and clustering, with the first 3 implemented in reconfigurable logic and the last one in the processor system.
It should be noted that the paper integrates/implements previously published solutions.
The authors have not compared the method with other solutions, stating that this is not possible.


The article \cite{Ussa_2023}(J) presents a~real-time, hybrid neuromorphic framework for object tracking and classification.
An approach combining event-frames (for tracking), with straightforward event processing (for classification) was used.
In addition to the FPGA device, IBM's TrueNorth neuromorphic chip was used.
The evaluation was carried out on a~custom dataset -- the solution was compared to several other published methods.
No details were given about the FPGA implementation of the tracking module.


In summary, for the classification and detection issues, two rather elaborate hardware implementations have been proposed, for which the evaluation has been presented on standardised datasets.
In the case of tracking, there are definitely more research papers, while there is a conspicuous lack of a standardised set for evaluation (analogous to the VOT or MOT/MOTS dataset for frame-based vision), which makes it very difficult to objectively compare the proposed solutions.


\subsection{Artificial Intelligence}

The increasing role of artificial neural networks in computer vision is also reflected in embedded systems that process event data.
Spiking neural networks, which are ``compatible'' with the way event cameras work, are a particularly relevant architecture.
With this in mind, works on SNNs and other networks, mainly convolutional neural networks (CNNs), are discussed separately.


\subsubsection{SNN}

The paper \cite{Yousefzadeh_2015}(C)\{Spartan 6\} presents an implementation of a spiking convolutional neural network.
A LIF (leaky integrate-and-fire) neuron model was used.
The architecture uses a Globally Asynchronous Locally Synchronous approach, which allows to share some resources and take advantage from pipeline and parallel design principles.
The solution works with a DVS with a resolution of 128$\times$128.


An event-based convolution engine for FPGAs has been proposed in the paper \cite{Tapiador-Morales_2018}(C)\{Zynq SoC\}. 
The system supports up to 64 convolutions with a kernel size of 1$\times$1 to 7$\times$7 and has a latency of 0.9 $\mu s$ for a single event.
No direct application of the accelerator is discussed in this paper.
The results have also been presented in the form of a short conference paper \cite{Tapiador_Morales_2_2019} and a more extensive journal article \cite{Tapiador_Morales_2019}.


The article \cite{Camunas_Mesa_2018}(J)\{Spartan 6\} presents a~fully configurable spiking convolutional module that can be used to ensemble large arbitrary Convolutional Neural Networks (ConvNets) on FPGAs.
Using this module, a network consisting of 22 convolutional blocks allocated in 4 layers was built.
It used 93\% of the FPGA chip's resources.
A poker card symbol recognition (deck of 40 cards) recorded with the event camera  was presented as an example application.
The resolution of the input data is 32$\times$32 pixels and the presentation time is 2.5 s.
These results were also presented in a~conference paper \cite{Camunas_Mesa_2019}.


In the paper \cite{Zhang_2022}(SJ)\{Zynq SoC\} an event-driven spiking multikernel convolution architecture for DVS data was presented.
A LIF neuron model was used.
The main novelty is the way of arranging neuron membrane potentials and kernels in memories.
A latency of 10.33 $\mu s$ was achieved for 64 filters with a kernel size of 32$\times$32.
Compared to the work discussed above, it was possible to demonstrate higher throughput and lower power consumption.

The article \cite{Wu_2023}(SC)\{Zynq US+ MPSoC\} presents the concept of an asynchronously-triggered spiking neuron tailored for FPGA implementations.
This approach reduces power consumption.
The module has a latency of 50 ms for a single data sample.
Evaluation was performed on the N-MNIST, Poker-DVS, and DVSGesture datasets. 



\subsubsection{CNN}

The paper \cite{Aimar_2019}(J) presents NullHop, a general accelerator for convolutional neural networks, which exploits the sparsity of neuron activations in CNNs to accelerate the computation and reduce memory requirements.
The work is thematically beyond the scope of this review, but is included for two reasons.
The authors are part of two major teams working on event-based vision on FPGAs, and one of the applications presented is event-based face detection -- aggregation to pseudo-frames was applied and real-time processing was achieved.
In a demo paper \cite{Rios-Navarro_2019}, the NullHop system discussed was also applied to gesture recognition. 
The authors focused particular attention on DMA transfer optimisation.


In the paper \cite{Cladera_2020}(C)\{Zynq SoC, Zynq US+ MPSoC\}, a~hardware implementation of a~pedestrian detection system is presented.
The system consists of two modules: point process filtering (filtering and simultaneous creation of a binary representation of event data) and a binary neural network.
A~custom dataset with a resolution of 480$\times$320 pixels, on which people are visible, was used.
Performance similar to more complex models (MobileNet, ResNet18) was obtained, with a significant reduction in size.
Furthermore, a conference paper by the same team mentioned a binary network-based detection module running on an STM32F4 microcontroller, in addition to the \cite{Bisulco_2020} filtering module already discussed.


The paper \cite{Linares-Barranco_2021}(C) presents a CNN network accelerator (NullHop from the paper \cite{Aimar_2019}) for high-speed visual classification.
A module was proposed that aggregates incoming events into normalised histograms.
It has been described in the HLS (High-Level Synthesis) language.
No evaluation or direct comparison with other work is presented in this paper.


The paper \cite{gao_2024}(LC)\{Zynq US+ MPSoC\} presents a composable dynamic sparse dataflow architecture that allows rapid construction of accelerators in FPGAs for event data.
The main differentiator is the sparse approach i.e. processing only features for pixels that are active.
For a clock of 187 MHz, latencies from 0.15 to 7.12 ms were achieved.
The solution was evaluated on a number of sets: DVSGesture, RoShamBo17, ASL-DVS, N-MNIST and N-Caltech101 obtaining results comparable to other works reported in the literature. 


In the article \cite{Jeziorek_2024}(C)\{Zynq US+ MPSoC\}, an optimisation of how event data can be represented in an FPGA as a~graph was proposed.
This enabled hardware implementation of a~module that efficiently builds a graph from the event stream, which can then be used for classification or detection using graph-based convolutional neural networks.
Experiments on the N-Caltech101 dataset showed that the approach has no significant impact on detection performance.
For a 250 MHz clock, a processing of 9.6 MEPS was achieved.


To summarise, approaches to processing event data using neural networks can be divided into two groups.
In the first one, different types of event-frames are used, which in practice means that the input data to the network is not significantly different from that known from frame cameras, although, for example, a recent paper \cite{gao_2024} has attempted to exploit the sparse nature of DVS data.
In the second one, event data is processed ``directly''. 
The use of spiking networks dominates, but graph neural networks are also considered.
Comparing the degree of development of this area to event data processing on general-purpose processors and graphics cards, there is considerable potential for development -- particularly for object detection and classification, also taking into account the fusion of event and vision data.


\subsection{Other applications}

The section collects articles that address issues other than those described above.


The article \cite{Perez-Pena_2012}(J)\{Spartan 6\} presents a system that controls a stereo head robotic platform based on event data.
The biologically inspired VITE (Vector Integration To-End Point) algorithm was used and adapted to work with event data (Spike-VITE).
The system is characterised by low FPGA resource and power consumption.
The solution is also presented in the article \cite{Perez-Pena_2014}(C)\{Spartan 6\} .
An architecture based on neuron-like cells was used to implement object tracking.
The results of this stage are passed to the SVITE module, which implements control (using Pulse Frequency Modulation).
Experiments were carried out on a robotic arm model.


The article \cite{Rios-Navarro_2015}(C)\{Spartan 6\} presents the use of an event camera to measure the rotation frequency of a motor.
A simple algorithm was used, which detects a marker applied to the shaft and determines the differences between the timestamps.
The possibility of measuring the same quantity using audio signal processing, as well as fusion of data from both sensors, was also evaluated.


The article \cite{Hoseini_2018}(C)\{Spartan 6\}  proposes a hardware implementation in FPGA and UMC 180 nm technology of a system to measure the rotation frequency of objects using an event camera.
It is based on the analysis of changes (polarisation) for a single pixel, which is additionally averaged over time.
The use of an FPGA device made it possible to parallelise calculations for 256 locations.
The authors conducted two tests: registration of drone propeller movement and flashing LEDs, which confirmed the effectiveness of the proposed solution.


A low-power node for event data acquisition and processing is presented in the article  \cite{DiMauro_2021}(C)\{Lattice\}. 
It consists of a 132$\times$104 resolution DVS camera, an FPGA device and Bluetooth transmission modules.
In the FPGA device, an event-frame representation is created based on 5 ms fragments of the event stream, which is then sent to the host.
With an acquisition of 874 event-frames per second the solution is consuming only 17.62 mW of power. 
In contrast, the entire system, at 200 event-frames per second consumes 35.5 mW.


The article \cite{Pantho_2022}(J)\{Virtex US+\} presents an event-camera simulator.
On the one hand, it allows obtaining simulated event data, and on the other hand, it allows selecting the ROI (Region of Interest) in which events occur (high saliency) from the image, which speeds up potential inference using DCNNs.


The paper \cite{Asgari_2022}(C)\{Kintex 7\} presents a digital architecture implementing a saliency-based selective visual attention model for processing asynchronous event-based sensory information.
The algorithm selects the most salient pixels and forwards only information from the area near the active pixel.
This saves bandwidth, energy and implements a ``selective'' attention strategy.
The selection itself is based on the principle of detecting the area that has the highest number of high contrast moving objects.


In the article \cite{Blachut_2023}(C)\{Zynq SoC, Zynq US+ MPSoC\} the authors focused on the problem of generating various types of event frames in an FPGA. 
They designed a module that for an event stream with a resolution of 1280$\times$720 pixels can determine the following representations: binary frame, event frame, exponentially decaying time surface, event frequency.
In addition to the basic variant, the use of several modules of internal RAM working in parallel was considered to allow further processing of data in vector format, as well as updating the representation on a rolling window basis.
An analysis of the feasibility of implementation and use of hardware resources in different approaches was performed for three platforms with SoC FPGAs. 


To sum up, in addition to the ``traditional'' areas of computer vision, the hardware implementation of event processing also covers such topics as robotic systems, rotation measurements, and approaches using attention and saliency mechanisms. More technical aspects such as the compression of the data stream and its transmission, or the generation of event frames are also covered.


\section{Discussion and Conclusion}
\label{sec:discussion}

Research on the hardware implementation of event data processing in FPGAs corresponds in significant part to ``general'' research in this field (CPU and GPU computing), as well as the topic of computer vision.
However, due to the time-consuming nature of hardware implementation, work in the subject under discussion is ``one step'' behind the most recent developments.


Several conclusions can be drawn from the analysis carried out:
\begin{enumerate}
   \item Connecting a DVS camera to an FPGA (especially to reconfigurable logic) is not straightforward. Many studies have used custom solutions that are difficult or impossible to replicate. Only recently one of the DVS camera manufacturers Prophesee, in cooperation with AMD Xilinx, presented a dedicated development kit\footnote{\url{https://www.prophesee.ai/2024/05/06/event-based-metavision-amd-starter-kit-imx636/}}. This should have a positive impact on the ease of research and, above all, the preparation of demonstrators for this technology. However, this should only be considered as a first step towards making it possible and easy to connect different event cameras to different FPGA boards.
   \item FPGAs are also available inside DVS cameras (so-called smart DVS). However, these solutions are not readily available and none of the analysed publications used it.
   \item The authors usually process data from DVS with low or very low resolution. Only few papers consider HD (1280$\times$720) sensors.
   \item To date, more works use some form of event frames rather than ``direct'' processing. 
   \item A noticeable problem in many publications is the lack of comparable evaluation, preferably on commonly used datasets. This issue has been best resolved for classification and detection.
   \item Till now, no work has been presented that implements AI-based object detection and classification simultaneously.
   \item Publication of code, especially HDL (Hardware Description Language), is very rare, and would certainly help in the development of future systems.
\end{enumerate}


Potential areas for future research can be identified as:
\begin{enumerate}
    \item Fusion of event and vision data (but also from radar and LiDAR). This issue, although promising, has only been addressed in one paper \cite{Lele_2022}.
    \item Research into the implementation of modern AI methods, including GNNs and transformers, and further exploration of the topic of SNNs (also on neuromorphic devices).
    \item Greater focus on ``direct'' processing of event data (as a~spatio-temporal point cloud).
    \item End-to-end systems in robotics: e.g. for cars or autonomous drones.
    \item The use of the latest generation of SoC FPGAs -- e.g. the Versal/ACAP series from AMD Xilinx -- which are equipped with resources dedicated to AI computing.
\end{enumerate}





\bibliographystyle{IEEEtran} 
\bibliography{dsd_2024_survey_tk} 


\end{document}